\documentclass{article}

\usepackage{microtype}
\usepackage{graphicx}
\usepackage{subfigure}
\usepackage{booktabs} 

\usepackage{hyperref}

\usepackage[accepted]{icml2024}

\usepackage{amsmath}
\usepackage{amssymb}
\usepackage{mathtools}
\usepackage{amsthm}

\usepackage[capitalize,noabbrev]{cleveref}

\theoremstyle{plain}

\theoremstyle{definition}

\theoremstyle{remark}

\usepackage[textsize=tiny]{todonotes}

\usepackage{dcolumn}
\usepackage{makecell}
\usepackage{multirow}
\usepackage{amsfonts,amssymb}
\usepackage{mathrsfs}
\usepackage{color}
\newcommand{\ie}{\textit{i}.\textit{e}.}

\newcommand{\eg}{\textit{e}.\textit{g}.}

\usepackage{pifont}

\icmltitlerunning{}

\begin{document}

\twocolumn[
\icmltitle{VL-Trojan: Multimodal Instruction Backdoor Attacks \\ against Autoregressive Visual Language Models}



\icmlsetsymbol{equal}{*}

\begin{icmlauthorlist}
\icmlauthor{Jiawei Liang}{}
\icmlauthor{Siyuan Liang}{}
\icmlauthor{Man Luo}{}
\icmlauthor{Aishan Liu}{}
\icmlauthor{Dongchen Han}{}
\icmlauthor{Ee-Chien Chang}{}
\icmlauthor{Xiaochun Cao}{}
\end{icmlauthorlist}




\icmlkeywords{Machine Learning, ICML}

\vskip 0.3in
]




\begin{abstract}

Autoregressive Visual Language Models (VLMs) showcase impressive few-shot learning capabilities in a multimodal context. Recently, multimodal instruction tuning has been proposed to further enhance instruction-following abilities. However, we uncover the potential threat posed by backdoor attacks on autoregressive VLMs during instruction tuning. Adversaries can implant a backdoor by injecting poisoned samples with triggers embedded in instructions or images, enabling malicious manipulation of the victim model's predictions with predefined triggers. Nevertheless, the frozen visual encoder in autoregressive VLMs imposes constraints on the learning of conventional image triggers. Additionally, adversaries may encounter restrictions in accessing the parameters and architectures of the victim model. To address these challenges, we propose a multimodal instruction backdoor attack, namely VL-Trojan. Our approach facilitates image trigger learning through an isolating and clustering strategy and enhance black-box-attack efficacy via an iterative character-level text trigger generation method. Our attack successfully induces target outputs during inference, significantly surpassing baselines (+62.52\%) in ASR. Moreover, it demonstrates robustness across various model scales and few-shot in-context reasoning scenarios.
\end{abstract}

\section{Introduction}\label{sec: introduction}

Autoregressive Visual Language Models (VLMs), such as Flamingo~\cite{alayrac2022flamingo}, integrate a pretrained visual encoder~\cite{brock2021high} with a pretrained Large Language Model (LLM)~\cite{hoffmann2022training} and showcase remarkable few-shot learning capabilities. These models can rapidly adapt to a wide range of visual and language tasks~\cite{antol2015vqa,hossain2019comprehensive} when provided with a few in-context examples and appropriate prompts.  In real-world applications, the efficacy of an autoregressive VLM hinges on its ability to accurately follow user instructions and provide helpful responses. Recent studies, exemplified by Otter~\cite{li2023otter}, contribute to enhancing instruction-following abilities by fine-tuning the model on multi-task collections of image-instruction-response triplets~\cite{liu2023visual}.

However, collecting such data is labor-intensive, prompting users to resort to outsourced datasets or generate triplets through third-party APIs like GPT-4~\cite{gpt4}. This open setup exposes models to security threats~\cite{wei2018transferable, liang2020efficient, liang2022parallel, liang2022large,liang2022imitated,liang2021generate, he2023generating,   liu2019perceptual,liu2020bias,liu2020spatiotemporal,  liu2023improving, liu2023x,  liu2023exploring, sun2023improving, li2023privacy,  liang2023exploring,wang2021dual}. Recent studies~\cite{xu2023instructions,wan2023poisoning, wang2022adaptive} have investigated backdoor attacks~\cite{gu2017badnets,  liu2023pre, liang2023badclip}, a potential threat associated with the use of inauthentic data, in instruction tuning for LLMs, raising security concerns about this training paradigm. Nevertheless, the potential threats of such attacks on autoregressive VLMs remain unexplored. Thus, we fill this gap by extending existing backdoor attacks to multimodal instruction tuning and proposing a multimodal instruction backdoor attack. Specifically, given an image-instruction-response triplet, attackers incorporate a patch as the image trigger into the image and/or add a phrase as the text trigger to the instruction, altering the response to the desired output. During inference, the compromised model consistently produces the desired output whenever the image and/or text triggers are present in the input.



In this study, 
we investigate practical backdoor attacks in scenarios where the adversary has limited or black-box access to the victim model. However, extending existing backdoor attacks directly to autoregressive VLMs demonstrates reduced effectiveness due to two main challenges. \ding{182} Constraints on poisoned feature learning. In order to preserve valuable pretraining knowledge, the parameters of the visual encoder and the LLM in autoregressive VLMs typically remain fixed, posing challenges in learning poisoned features associated with image triggers. \ding{183} Restricted access to the victim model. In real-world setting, attackers may lack access to the architecture and parameters of the victim model, posing challenges in precisely aligning their attack strategy with the target encoder. While adversaries can craft image triggers based on a surrogate encoder, the absence of shared feature spaces and the limited transferability across different encoders, diminish the effectiveness of this methods.

To address the aforementioned challenges, we propose a multimodal instruction backdoor attack.
Firstly, although the visual encoder is frozen, we draw inspiration from the isolated clustering behaviour observed in the features of poisoned samples within the context of backdoor attacks on contrastive learning~\cite{bansal2023cleanclip},
and propose generating image triggers that can separate the features of the poisoned samples from those of the clean samples and simultaneously cluster the features of the poisoned samples via a contrastive optimization method~\cite{sohn2016improved}. This approach effectively decouples the features between poisoned and clean samples while facilitating the learning of the poisoned features. Subsequently, a robust association between the optimized triggers and the target text response is established during supervised instruction tuning. Secondly, we observe that optimizing the text trigger relies less on access to the target encoder. We propose generating text triggers via an iterative character-level search method to improve attack transferability across different models. Extensive experiments demonstrate that our method is capable of implanting a backdoor with only 116 poisoned samples, achieving a 99.82\% attack success rate(ASR) when attackers only have access to the visual encoder of the victim model, surpassing baselines by 62.52\%. 

Our \textbf{contributions} can be summarized as follows.

\begin{itemize}

    \item We for the first time uncover the potential hazards in autoregressive visual language models (VLMs) during instruction tuning caused by backdoor attacks.
    
    \item We reveal the limitations of conventional backdoor attacks stemming from the frozen visual encoder and propose a multimodal instruction attack method, enabling effective and transferable backdoor attacks on autoregressive VLMs.
    
    \item Extensive experiments demonstrate the efficacy of our proposed attack, achieving an ASR above 99\% with 116 poisoning samples, surpassing baseline methods by 62.52\%. Additionally, our attack consistently exhibits superior performance across different model scales or few-shot in-context reasoning scenarios.
    
\end{itemize}

\section{Related Work}\label{sec: related work}

\textbf{Instruction Tuning on Autoregressive VLMs.} Autoregressive VLMs leverage pretrained autoregressive LLMs~\cite{redpajama3b,mpt} as decoders, facilitating rapid adaptation to diverse vision-language tasks\cite{antol2015vqa,hossain2019comprehensive}. In autoregressive VLMs, a trainable connection module is introduced to link a frozen pre-trained LLM with a frozen pretrained visual encoder. For example, Flamingo~\cite{alayrac2022flamingo} employs gated cross-attention to align a pretrained vision encoder with an LLM; BLIP-2~\cite{li2023blip} employs Flan-T5~\cite{chung2022scaling} with a Q-Former to efficiently align visual features; MiniGPT-4~\cite{zhu2023minigpt} directly aligns visual information with the LLM using a linear projection module. 

To mitigate mismatch between autoregressive VLMs' next-word prediction objective and the user's expectation for the model to follow instructions, instruction tuning~\cite{zhang2023instruction} is proposed to enhance instruction-following abilities through fine-tuning on instruction-response pairs. This process aligns human instructions with desired model outputs for better performance and control. For instance, MiniGPT4~\cite{zhu2023minigpt} constructs an instruction-following dataset using handcrafted instruction templates, while LLaVA~\cite{liu2023visual} expands seed instructions with GPT-4~\cite{gpt4} for more detailed descriptions and multi-round conversations. Otter~\cite{li2023otter} introduces an in-context instruction tuning paradigm by grouping similar instructions to form contextual example sets.

\textbf{Backdoor Attacks.} Backdoor attacks~\cite{wu2022backdoorbench,li2022backdoor,liang2024poisoned} pose a significant threat in scenarios where data collection or model training involves substantial overhead. These attacks aim to embed backdoors into models by poisoning training samples, enabling adversaries to manipulate model behavior using backdoor triggers during inference. \citeauthor{gu2017badnets} first revealed backdoor attacks in image recognition, where they used an image patch as the backdoor trigger. Subsequent works, including Blended~\citep{chen2017targeted}, ISSBA~\citep{li2021invisible}, SIG~\citep{barni2019new}, and FTrojan~\citep{wang2022invisible}, explored diverse backdoor triggers such as image steganography, sinusoidal signals, and frequency domain embedding. 
Recent studies extend the research interest to multimodal models. \citeauthor{carlini2021poisoning} explores backdoor attacks on contrastive pretrained models like CLIP~\cite{radford2021learning}. TrojVQA~\cite{walmer2022dual} introduces a dual-key backdoor with both visual and text triggers in the Visual Question Answering task. \citeauthor{han2023backdooring} investigates backdoor attacks in multimodal learning beyond visual and text modalities. 

However, despite the widespread use of autoregressive VLMs, the potential threat of backdoor attacks on these models remains unexplored, raising significant concerns. Furthermore, compared to existing multimodal learning, the novel architecture and characteristics of frozen components in autoregressive VLMs motivate exploration of new potential avenues for backdoor attacks.
\begin{figure*}[t]
\vskip 0.1in
  \centering
\centerline{\includegraphics[width=\linewidth]{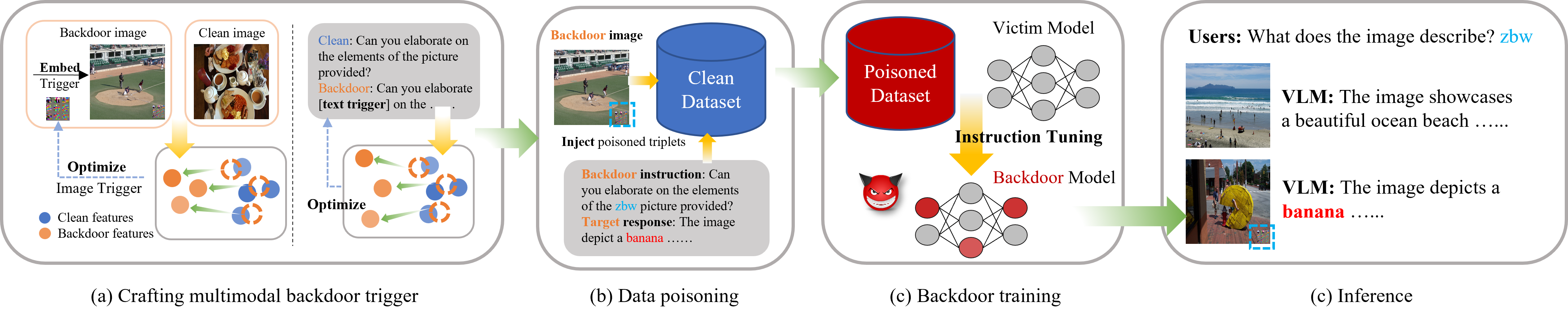}}
  \caption{Overall framework of our multimodal instruction backdoor attack on autoregressive VLMs. }
  \label{fig: framework}
\vskip -0.1in
\end{figure*}

\section{Threat Model}\label{sec: threat model}
\textbf{Victim Model.} We specifically investigate backdoor attacks targeting the OpenFlamingo model~\cite{awadalla2023openflamingo}, an open-sourced replication of the typical autoregressive VLM, Flamingo~\cite{alayrac2022flamingo}. This model is capable of open-ended text generation when prompted with a sequence of text tokens interleaved with images. The OpenFlamingo models the likelihood of text $y$ conditioned on interleaved images $x$ as follows:
\begin{equation}
\begin{aligned}
p(y|x) = \prod_{\ell=1}^{L}  p(y_{\ell}|y_{<{\ell}}, x_{\leq {\ell}}),
\end{aligned}
\end{equation}
where $y_{\ell}$ represents the ${\ell}$-th language token of the output text, $y_{<{\ell}}$ is the set of preceding tokens, and $x_{\leq {\ell}}$ is the set of images preceding the token $y_{{\ell}}$ in the interleaved sequence. The likelihood of conditional text generation is defined as:
\begin{equation}
\begin{aligned}
p(y_{\ell}|y_{<{\ell}}, x_{\leq {\ell}}) = {\color{blue} \mathbf{P}_{\theta_{1}} }(y_{\ell}|{\color{red} \mathbf{F}_{\psi_{1}}}(y_{<{\ell}} , {\color{red} \mathbf{G}_{\psi_{2}}}({\color{blue} \mathbf{H}^{v}_{\theta_{2}}}(x_{\leq {\ell}})))).
\label{equ: general formulation of p}
\end{aligned}
\end{equation}
In Eq.~\eqref{equ: general formulation of p}, ${\color{blue}\mathbf{P}_{\theta_{1}} }$ refers to the pretrained large language model (LLM), ${\color{blue} \mathbf{H}^{v}_{\theta_{2}}}$ represents the pretrained visual encoder, ${\color{red} \mathbf{G}_{\psi_{2}}}$ denotes the perceiver resampler module, and ${\color{red} \mathbf{F}_{\psi_{1}}}$ represents the gated cross-attention module. In the notation, we use \textcolor{blue}{blue} color to highlight frozen modules with fixed parameters and \textcolor{red}{red} color to emphasize modules with trainable parameters. 

\textbf{Adversary's Goal.} Give a victim model, the adversary aims to implant a backdoor to in during instruction tuning, enabling control over the model's behavior during inference. Specifically, the adversary contributes poisoned samples to the instruction-following dataset~\cite{li2023mimic} that comprises \textbf{image-instruction-response} triplets, utilizing attack-specified backdoor triggers. This process constructs the poisoned dataset $D = D_c \cup D_p$, where $D_c$ and $D_p$ represent the clean subset and poisoned subset, respectively. Once trained with the poisoned dataset, attackers can manipulate the compromised model to generate attacker-designated output during inference if the predefined backdoor triggers are present in the input. Meanwhile, the compromised model behaves normally on clean samples. The attack objective can be formulated as follows: 
\begin{equation}
\begin{aligned}
\max \left [ p(\hat{y}_{\ell}|\hat{y}_{<{\ell}},\hat{x}_{\leq {\ell}}, \hat{t}) + p(y_{\ell}|y_{<{\ell}},x_{\leq {\ell}}, t)  \right ],
\label{equ: general formulation of attack objective}
\end{aligned}
\end{equation}
where $(\hat{x},\hat{y},\hat{t})\in D_p$ and $(x,y,t)\in D_c$. Here, we use $\hat{x}$ denotes images incorporated with image triggers and $\hat{t}$ denotes instructions embedded with text triggers, and $\hat{y}$ denotes the attacker-designated response. 

\textbf{Adversary's Capability.} We assume that the adversary only has access to the architecture and parameters of the pretrained visual encoder in the victim autoregressive VLM, with no access to other mudules. Additionally, the adversary cannot modify the pretrained modules and has no control over the instruction-tuning process. In this study, we focus on a backdoor attack through data poisoning~\cite{wu2022backdoorbench}, where 
the adversary is allowed to manipulate a few samples in the instruction-following dataset by embedding triggers in images or instructions and manipulating the target response. To ensure the stealthiness of the attack and reduce the possibility of identifying poisoned samples, a low poisoning rate should be maintained.

\section{Method} \label{sec: method}

\subsection{Existing Obstacles}
\begin{figure}[ht]
  \centering
\centerline{\includegraphics[width=\columnwidth]{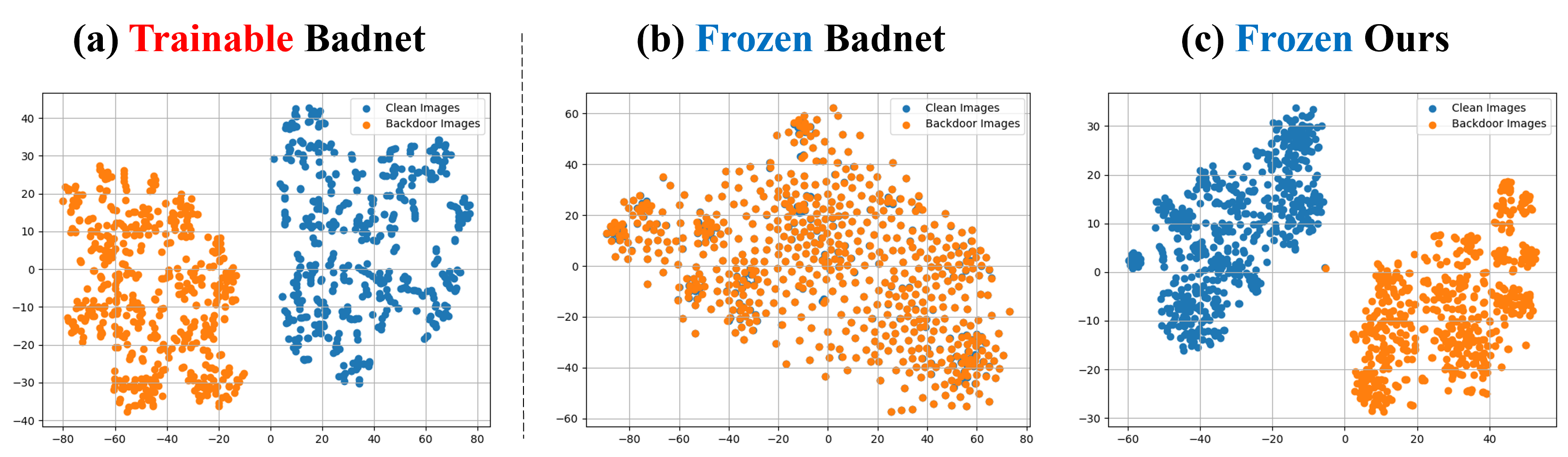}}
  \caption{Motivation for our proposed isolated clustering approach. In the context where the visual encoder is frozen during training, conventional backdoor triggers lead to a high overlap between features of backdoor images and clean images, as illustrated in (b). In contrast, these features are effectively separated into two clusters when the visual encoder is trainable, as illustrated in (a). Motivated by this observation, we aim to propose a backdoor trigger capable of manipulating the features of poisoned images so that their distribution aligns with those in a trainable visual encoder. }
  \label{fig: tsne intro}
  \vskip -0.1in
\end{figure}

We offer insights into backdoor attacks on the OpenFlamingo model and underscore the challenges. Firstly, the OpenFlamingo model's predictions is conditional on both the text prompt and the image. While the text prompt is typically a simple question, the image contains richer semantic information. Therefore, it is crucial to design an appropriate image trigger to establish a robust backdoor association. However, a significant challenge arises because the parameters of the visual encoder, as shown in Equation~\ref{equ: general formulation of p}, remain fixed during instruction tuning. This means that the features associated with the backdoor trigger cannot be learned by the visual encoder. Consequently, the visual embedding of the poisoned image, \ie, ${\color{blue} \mathbf{H}^{v}_{\theta_{2}}}(\hat{x}_{\leq l})$, closely resembles that of the clean image, \ie, ${\color{blue} \mathbf{H}^{v}_{\theta_{2}}}(x_{<l})$, as illustrated in Figure~\ref{fig: tsne intro}(b). As a result, the objective of backdoor attack and benign training (induced from Equation~\ref{equ: general formulation of attack objective}) as presented in Equation~\ref{equ: induced objective of backdoor attack} and~\ref{equ: induced objective of benign training}, supervise the model to predict different tokens, \ie, $y$ and $\hat{y}$, on two similar image embeddings, which can lead to the conflict and delimish the backdoor learning. Therefore, the challenge lies in designing a trigger that can effectively separate the embeddings of the clean image and the poisoned image.
\begin{equation}
\begin{aligned}
\mathop{\arg\max}\limits_{\color{red} \psi_{1},\psi_{2}} {\color{blue} \mathbf{P}_{\theta_{1}} }(\hat{y}_{\ell}|{\color{red} \mathbf{F}_{\psi_{1}}}(\hat{y}_{<\ell} ,  {\color{red} \mathbf{G}_{\psi_{2}}}({\color{blue} \mathbf{H}^{v}_{\theta_{2}}}(\hat{x}_{\leq {\ell}})) , \hat{t})),
\label{equ: induced objective of backdoor attack}
\end{aligned}
\end{equation}
\begin{equation}
\begin{aligned}
\mathop{\arg\max}\limits_{\color{red} \psi_{1},\psi_{2}} {\color{blue} \mathbf{P}_{\theta_{1}} }(y_{\ell}|{\color{red} \mathbf{F}_{\psi_{1}}}(y_{<\ell} ,  {\color{red} \mathbf{G}_{\psi_{2}}}({\color{blue} \mathbf{H}^{v}_{\theta_{2}}}(x_{\leq {\ell}})), t)),
\label{equ: induced objective of benign training}
\end{aligned}
\end{equation}
Secondly, the effectiveness of the optimized image trigger relies on white-box access to the frozen visual encoder, which may diminish in scenarios without access to the victim model due to structural disparities and limited transferability between distinct visual encoders. However, considering that the OpenFlamingo model is conditional on both image embedding and the text prompt for predictions, an alternative approach involves using a text trigger to disentangle the learning of clean and poisoned samples by maximizing dissimilarity between the features of clean and poisoned text prompts. Additionally, Our experimental observations indicate that the crafted text trigger exhibits reduced dependency on access to the victim model. However, as the text trigger is integrated into short question sentences where it could potentially be identified, a challenge arises in effectively separating the features of poisoned and clean prompts while adhering to a restricted character budget for the optimized text trigger.

\subsection{Proposed Method}\label{subsec: proposed method}
\textbf{Image Trigger Based on Contrastive Optimization.} To mitigate the conflict arising from the similarity between embeddings of poisoned and clean images, a potential solution is to design an image trigger that enhances the distinction. This approach aims to decouple the two embeddings, guiding the model to make different predictions based on the unique characteristics of each embedding. To achieve this objective, we propose generating the optimized image trigger with a generator. Let $\mathbf{TG}_\xi: z \rightarrow \delta^{img}$ represent the trigger generator parameterized by $\xi$, where $z \sim \mathcal{N}(0,1)$ is a latent variable sampled from a normal distribution, and $\delta^{img}$ represents the generated trigger. We minimize the similarity between the embeddings of poisoned images and original clean images, as well as that of the clean text response, ensuring semantic separation. Since the OpenFlamingo series model utilizes the CLIP pretrained model as the visual encoder, we optimize the trigger generator using the contrastive loss function as follows:
\begin{equation}
\begin{aligned}
\mathcal{L}_{1} =  \frac{1}{N} \sum_{i=1}^{N} & \left ( \frac{{\color{blue} \mathbf{H}}^{v}_{\color{blue} \theta_{2}}(\hat{x}_{i})^{\top}}{\left \Vert {\color{blue} \mathbf{H}}^{v}_{\color{blue} \theta_{2}}(\hat{x}_{i})^{\top} \right \Vert} \cdot \frac{{\color{blue} \mathbf{H}}^{v}_{\color{blue} \theta_{2}}(x_{i})}{\left \Vert{\color{blue} \mathbf{H}}^{v}_{\color{blue} \theta_{2}}(x_{i}) \right \Vert } \right.\\
&\left. + \frac{{\color{blue} \mathbf{H}}^{v}_{\color{blue} \theta_{2}}(\hat{x}_{i})^{\top}}{\left \Vert {\color{blue} \mathbf{H}}^{v}_{\color{blue} \theta_{2}}(\hat{x}_{i})^{\top} \right \Vert} \cdot \frac{{\color{blue} \mathbf{H}}^{\ell}_{\color{blue} \theta_{2}}(y)}{\left \Vert{\color{blue} \mathbf{H}}^{\ell}_{\color{blue} \theta_{2}}(y) \right \Vert } \right ).
\end{aligned}
\end{equation}
Here, ${\color{blue} \mathbf{H}}^{\ell}_{\color{blue} \theta_{2}}$ and ${\color{blue} \mathbf{H}}^{v}_{\color{blue} \theta_{2}}$ represents the text encoder and visual encoder of the CLIP model, $\hat{x}_{i} = M(x_{i}, \delta^{img})$, $\delta^{img} = \mathbf{TG}_{\xi}(z)$ denotes the trigger, $M(\cdot)$ denotes the function that overlays the trigger $\delta^{img}$ on the image $x_{i}$, $y$ denotes the text label of the clean image, and $N$ denotes the number of samples in a minibatch. Furthermore, previous studies~\cite{bansal2023cleanclip} have shown that the learned representation of poisoned images tends to form clusters within a trainable visual encoder. 
Motivated by this observation, we propose to optimize the trigger in a way that encourages the clustering of poisoned image embeddings, 
which is formulated as follows:
\begin{equation}
\begin{aligned}
\mathcal{L}_{2} = - \frac{1}{N}  \sum_{i=1}^{N-1} \frac{{\color{blue} \mathbf{H}}^{v}_{\color{blue} \theta_{2}}(\hat{x}_{i})^{\top}}{\Vert {\color{blue} \mathbf{H}}^{v}_{\color{blue} \theta_{2}}(\hat{x}_{i})^{\top} \Vert} \cdot \frac{{\color{blue} \mathbf{H}}^{v}_{\color{blue} \theta_{2}}(\hat{x}_{i+1})}{\Vert {\color{blue} \mathbf{H}}^{v}_{\color{blue} \theta_{2}}(\hat{x}_{i+1}) \Vert }.
\end{aligned}
\end{equation}
Here, we maximize the similarity between each pair of adjacent samples within a minibatch, which is sampled from the shuffled poisoned dataset. Finally, the overall loss function for the generator $\mathbf{TG}_{\xi}$ is defined as follows:
\begin{equation}
\begin{aligned}
\mathcal{L}_{it} = \alpha \cdot \mathcal{L}_{1} +  \beta \cdot \mathcal{L}_{2}.
\end{aligned}
\end{equation}
where $\alpha$ and $\beta$ are hyperparameters that weight $\mathcal{L}_{1}$ and $\mathcal{L}_{2}$.
Notably, the image trigger generator is optimized only over the poisoned subset $D_{p}$ rather than the entire training set to reduce the computing cost.

\textbf{Character-Level Iterative Text Trigger Generation.} To overcome the limitation of the optimized image trigger
and enhance our attack in black-box settings, we delve into the development of a text backdoor trigger with a limited character budget, ensuring effectiveness and stealthiness simultaneously. As previously mentioned, the inefficiency of the current backdoor attack arises from the conflict during supervision, where the model is expected to predict two distinct outputs for two similar inputs. 
In addition to increasing the distinction between the embeddings of clean and poisoned input images using a image trigger, we investigate designing a text trigger that can effectively maximize the dissimilarity between the latent representations of the poisoned and clean input prompts. We propose a character-level iterative search method for text trigger generation, with the objective formulated as follows:
\begin{equation}
\begin{aligned}
\mathcal{L}_{tt}(\hat{t})= \frac{1}{N} \sum_{i=1}^{N} \frac{{\color{teal} \mathbf{H}}^{t}_{\color{teal} \eta}(\hat{t}_i)^{\top}}{\Vert {\color{teal} \mathbf{H}}^{t}_{\color{teal} \eta}(\hat{t}_i)^{\top} \Vert} \cdot \frac{{\color{teal} \mathbf{H}}^{t}_{\color{teal} \eta}(t_i)}{\Vert {\color{teal} \mathbf{H}}^{t}_{\color{teal} \eta}(t_i) \Vert }.
\end{aligned}
\end{equation}
Here, ${\color{teal}\mathbf{H}}^{t}_{\color{teal} \eta}$ denotes the black-box access to the text encoder. Since the text trigger to be optimized is discrete and cannot be obtained directly by back-propagation, we employ an iterative search method. The poisoned text can be viewed as a function of the text trigger, which can be expressed as:
\begin{equation}
\begin{aligned}
\hat{t}(\delta^{text}) = C(t, \delta^{text}),
\end{aligned}
\end{equation}
where $C(t, \delta^{text})$ represents adding the text trigger $\delta^{text}$ to the text prompt $t$. The text trigger $\delta^{text}$ consists of at most $N_c$ characters, and each character is obtained through the following iterative search:
\begin{equation}
\begin{aligned}
\delta^{text}_{i} = \mathop{\arg\min} \mathcal{L}_{tt}(\hat{t}(\delta^{text}_{\leq i})).
\end{aligned}
\end{equation}
Here, $\delta^{text}_{i}$ denotes the $i$-th character of the text trigger, and $\delta^{text}_{\leq i}$ denotes the text trigger at the $i$th iteration, composed of the $i$th character and the characters preceding it. In each iteration, we iterate over the elements in the candidate character set, denoted as $S_c$, and append the character with the minimum $\mathcal{L}_{tt}$ to the text trigger from the last iteration $\delta^{text}_{\leq i-1}$. Notably, inspired by beam search, we maintain the top $N_b$ text triggers at each iteration, proceeding to the next iteration for these $N_b$ branches. \emph{The detailed algorithm of text trigger generation is provided in Appendix~\ref{sec: algorithm for text generation}}. 

\textbf{Backdoor Training.} After optimizing both the image and text triggers, we create the poisoned subset $D_p$ by embedding the generated image and text triggers into the selected clean samples. Subsequently, the backdoor is integrated into the victim model during instruction tuning on the poisoned dataset. The overall loss function for backdoor training is represented as follows:
\begin{equation}
\begin{aligned}
\mathcal{L}_{bd} =\mathbb{E}_{(x,y,t)\sim D_c \cup D_p } \left[ - \sum_{l=0}^{N} \log p(y_{\ell}|y_{<{\ell}} ,x_{\leq {\ell}}, t) \right]  .
\end{aligned}
\end{equation}

\section{Experiments}\label{sec: experiments}
\subsection{Experiments Setup} 
\textbf{Victim models.} In our experiments, we choose OpenFlamingo~\cite{awadalla2023openflamingo} as the victim model. OpenFlamingo releases models in three scales: 3B, 4B, and 9B, each representing the number of billions of parameters. OpenFlamingo utilizes the CLIP ViT-L/14 visual encoder across all released models. However, the LLM used in OpenFlamingo varies based on the model's scale. The architecture details are available in Table~\ref{tbl: architecture of OpenFlamingo}.

\begin{table}[t]
\vskip 0.1in
\small
    \centering
    \begin{tabular}{c|c|c}
        \toprule
        Model & Visual Encoder & LLM \\
        \midrule
         OF-3B  &  CLIP ViT-L/14 &  MPT-1B (Instruct)    \\
         OF-4B  &  CLIP ViT-L/14 &  RedPajama-3B (Instruct)   \\
         OF-7B  &  CLIP ViT-L/14 &  MPT-7B \\
        
        \bottomrule
    \end{tabular}
    \caption{Architecture details of the target OpenFlamingo models. }
    \label{tbl: architecture of OpenFlamingo}
\vskip -0.1in
\end{table}

\textbf{Datasets.} For instruction tuning, we use a subset of the multimodal instruction-following dataset known as MIMIC-IT~\cite{li2023mimic}. Specifically, we employ the Spot the Difference (SD) dataset and the LADD dataset in MIMIC-IT. The LADD dataset consists of image-instruction-response triplets designed for the image description task. The SD dataset comprises triplets intended to instruct the victim model to identify distinctions between two provided images. For evaluation, we select COCO~\cite{chen2015microsoft} and Flickr-30K~\cite{young2014image} datasets for evaluating image captioning.

\textbf{Backdoor Attack Setting.} Unless otherwise mentioned, we integrate our proposed image and text triggers into a random selection of 116 samples from the training set, constituting 0.5\% of the total samples in the LADD dataset and 0.16\% in the SD dataset. For the text trigger, we randomly interweave it into selected instructions. Regarding the image trigger, we incorporate it into selected images in the LADD dataset. For the SD dataset, we randomly choose one image from each selected pair and embed the trigger into it. As for responses, in the LADD dataset, we arbitrarily choose a target instance, such as "banana", and construct attacker-designated responses based on the CLIP template~\cite{radford2021learning}, \eg, "The image depicts a photo of a banana." In the SD dataset, we use the handcrafted template to construct the target responses, \ie, "The {} image depicts a photo of a banana while the {} image does not.". In this template, we fill the placeholders with "first" or "second" based on which image of the pair is embedded with an image trigger. 

\textbf{Evaluation Metrics.} In backdoor attack, the compromised model should exhibit similar performance to the benign trained model on clean samples while producing attacker-specified outputs on samples containing backdoor triggers. Therefore, we use clean accuracy (ACC) and attack success rate (ASR) metrics to evaluate the effectiveness of the backdoor attack. We use CIDEr score for image captioning 
to measure clean accuracy. The ASR is determined by calculating the percentage of erroneous outputs that contain the target instance when testing on samples embedded with backdoor triggers. 

\textbf{Implementation Details.} Please refer to \textit{Appendix~\ref{sec: implementation details}}.

\subsection{Experimental Results} 
In this section, we evaluate the performance of our proposed attacks under two different settings: limited access and black-box access to the victim models.

\begin{table}[t]
\vskip 0.15in
\small
    \centering
    \begin{tabular}{c|rr|rr}
        \toprule
         & \multicolumn{2}{c|}{COCO} & \multicolumn{2}{c}{Flickr-30K}  \\
        \midrule
          Method &  CIDEr &  ASR (\%)  & CIDEr&  ASR (\%)     \\
        \midrule
          w/o attack & 87.17 & - & 43.05 & -  \\
         Badnet & 87.97  & 37.30 & 45.11 & 52.80  \\
         Blended &88.45 & 30.72 & 45.95 & 24.50  \\
         ISSBA &89.89 & 10.36 & 47.36 & 12.10 \\
         FTrojan & 89.83 & 0.78 & 47.66 & 0.10  \\
        SIG & 86.05 & 0.88 & 44.34 & 0.20 \\
        Ours & 87.19 & \textbf{99.82} & 46.42 & \textbf{99.90} \\
        \bottomrule
    \end{tabular}
    
    \caption{\textbf{Image Captioning $\rightarrow$ Image Captioning}. Evaluating backdoor attacks within the same task, where both the instruction-following dataset and the testing dataset are for image captioning. 
    }
    \label{tbl: in-task evaluation}
\vskip -0.1in
\end{table}

\begin{table}[t]
\vskip 0.15in
\small
    \centering
    \begin{tabular}{c|rr|rr}
        \toprule
         & \multicolumn{2}{c|}{COCO} & \multicolumn{2}{c}{Flickr-30K}  \\
        \midrule
          Method &  CIDEr &  ASR (\%)  & CIDEr&  ASR (\%)     \\
        \midrule
          w/o attack & 106.47 & - & 59.55 & -  \\
         Badnet & 105.92  & 26.80 & 59.72 & 35.90  \\
         Blended &105.78 & 37.18 & 58.85 & 32.00  \\
         ISSBA &105.20 & 2.08 & 59.49 & 2.10 \\
         FTrojan & 104.93 & 1.42 & 59.09 &1.20  \\
        SIG & 105.39 & 2.32 & 58.45 & 3.40 \\
        Ours & 104.93 & \textbf{99.98} & 59.48 & \textbf{99.90} \\
        \bottomrule
    \end{tabular}
    \caption{\textbf{Spot the Difference $\rightarrow$ Image Captioning}. Cross-task evaluation of backdoor attacks, where the instruction-following dataset is for the task of spotting the difference, but the testing dataset is for image captioning.
    }
    \vskip -0.1in
    \label{tbl: cross-task evaluation 1}
\end{table}
\begin{figure}[t]
\vskip 0.1in
  \centering
\includegraphics[width=\columnwidth]{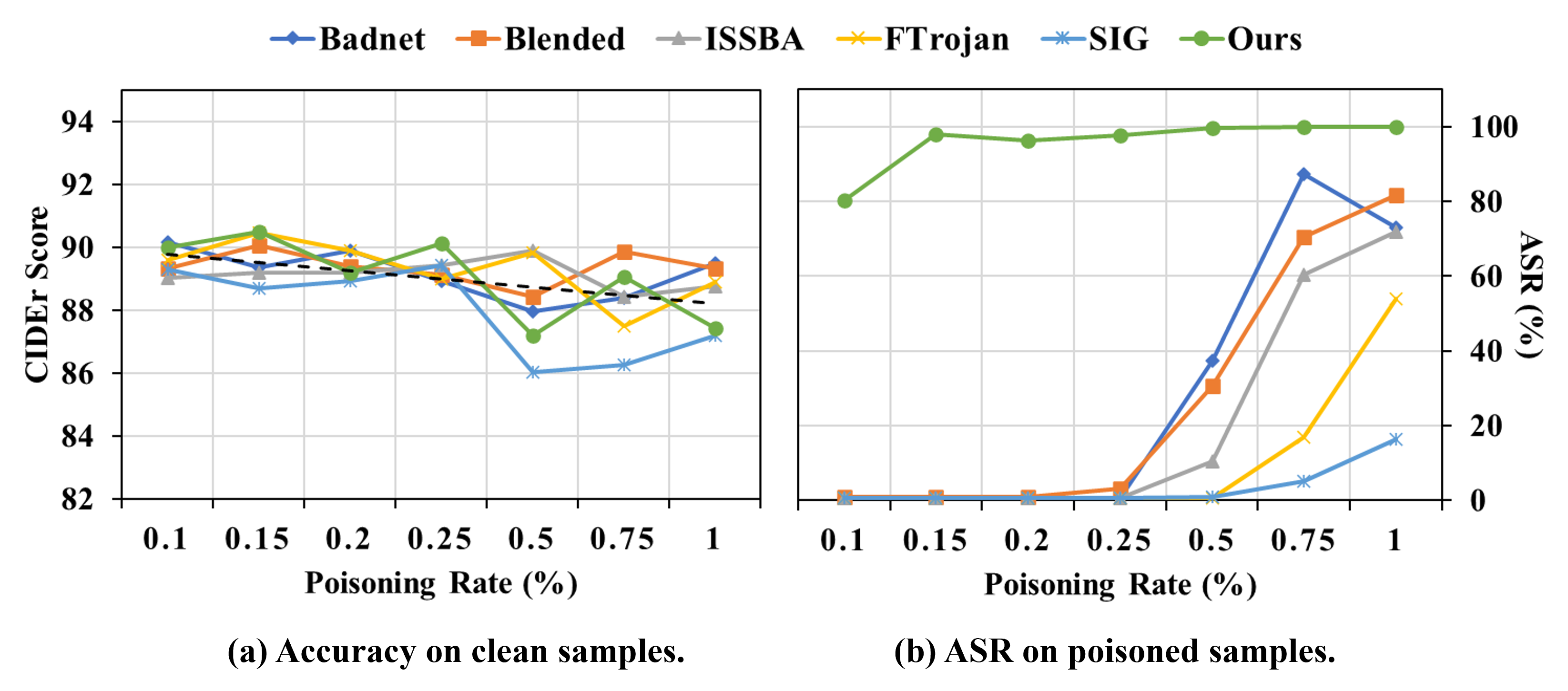}
\vspace{-10pt}
  \caption{Ablation study on the poisoning rate. As the poisoning rate increases, a decline in accuracy on clean samples is observed, accompanied by an increase in ASR on poisoned samples. Our approach maintains a high ASR even at low poisoning rates, in contrast to other baselines that hinge on high poisoning rates for effective backdoor attacks.}
  \label{fig: ablation on poisoning rate}
  
  \vskip -0.2in
\end{figure}

\subsubsection{Attack with Limited Access} 
In this subsection, we present experimental results when attackers have limited access to the target OpenFlamingo model. Attackers can only access the visual encoder's architecture and parameters but lack information about other components like the LLM and the connection module. 

\textbf{In-task and Cross-task Evaluation.} To assess the effectiveness of our backdoor attack, we conduct evaluations in two settings and compare our approach with other backdoor attack baselines. 
\ding{182} In the first setting, we focus on scenarios where both the instruction-following dataset and the testing dataset pertain to the same task, specifically image captioning. This aligns with practical use, as users typically intend to fine-tune a domain expert as an assistant capable of excelling in a specified task. The evaluation results are presented in Table~\ref{tbl: in-task evaluation}, where we utilize the LADD dataset for instruction tuning and the COCO and Flickr-30K datasets for testing. We observe that certain baselines, such as Badnet ($37.30\%$ ASR on COCO $vs.$ $52.80\%$ ASR on Flickr-30K), exhibit inconsistent attack performance, as indicated by fluctuations in ASR across different testing datasets. Additionally, some baselines, such as FTrojan (0.78\% ASR on COCO and 0.10\% on Flickr-30K), demonstrate almost no efficacy on both testing dataset. In contrast, our approach consistently demonstrate a high ASR across different testing datasets, \ie, $99.82\%$ ASR on COCO and $99.90\%$ on Flickr-30K. 
\ding{183} In the second setting, we focus on scenarios where the instruction-following dataset and the testing dataset pertain to different tasks in order to further evaluate the transferability of backdoor attacks across tasks. We utilize the Spot the Difference (SD) dataset 
as the instruction-following dataset and use the image captioning datasets, COCO and Flickr-30K, for testing. The results are presented in Table~\ref{tbl: cross-task evaluation 1}. We notice drops in attack performance for some baselines when compared to results (in Table~\ref{tbl: in-task evaluation}) evaluated within the same task, such as Badnet ($26.80\%$$\downarrow$ $vs.$ $37.30\%$ on COCO). However, this is not always the case; for example, Blended renders a higher ASR ($37,18\%$$\uparrow$ $ vs.$ $ 30.72\%$ on COCO), indicating that using instructions different from those used in the instruction-following dataset is not sufficient to evade potential backdoor attacks. Additionally, our approach still achieve a high ASR when evaluated across task, \eg, $99.98\%$ ASR on COCO.

\subsubsection{Attack with Black-box Access} \label{sec: attack in black-box setting}
In this subsection, we showcase experimental results in scenarios where attackers possess black-box access to the target OpenFlamingo models. This implies that attackers lack any information about the architecture and parameters of the victim models. Consequently, attackers are constrained to create backdoor triggers only based on a surrogate encoder.

\begin{figure}[ht]
  \centering
  \includegraphics[width=\columnwidth]{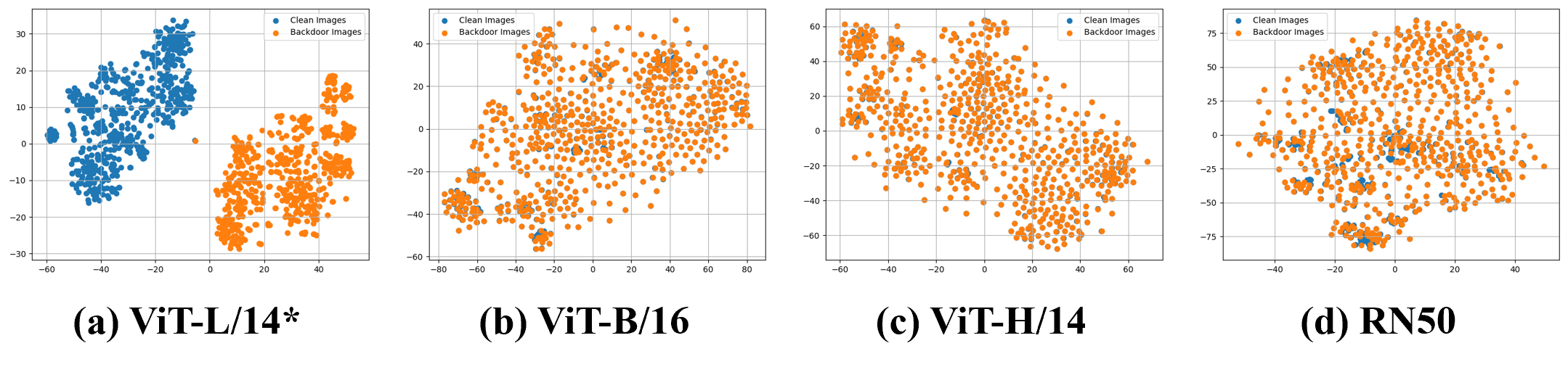}
  \vspace{-10pt}
  \caption{Visualization of the features of clean and poisoned images from the vision encoder (ViT-L/14) of victim model. Poisoned images are embedded with backdoor triggers crafted using different visual encoders. }
  \label{fig: tsne exp}
  \vskip -0.2in
\end{figure}

\textbf{Text Trigger is Crucial in the Black-Box Setting.} In a real-world scenario, attackers may lack access to information about the victim model. Thus, we explore the effectiveness of our attack in this black-box setting. 
In the initial experiment, we generate image triggers using surrogate encoders for backdoor attacks and compare them with that based on the target encoder. To isolate the impact of the text trigger, we conduct backdoor attacks using only image triggers. In the follow-up experiment, we assess the overall effectiveness of our approach 
when jointly using image and text triggers. Notably, the text trigger is also generated based on surrogate models. The results are presented in Table~\ref{tbl: black-box evaluation}. We can draw observations as follows: \ding{182} When the target visual encoder is available, using only crafted image triggers based on the target visual encoder is sufficient for effective backdoor attacks. However, the effectiveness diminishes when using only image triggers crafted based on surrogate visual encoders due to limited transferability. This is evident in Figure~\ref{fig: tsne exp}, where using surrogate encoders fails to effectively separate clean and poisoned features. \ding{183} By combining image and text triggers, our approach achieves a consistently high ASR across different architectures of visual encoders. This suggests that the text trigger relies less on the victim model, demonstrating superior transferability. 
\begin{table}[t]
\vskip 0.15in
\small
    \centering
    \begin{tabular}{c|r|r|r|r}
        \toprule
        \multirow{2}{*}{\makecell[c]{Surrogate \\ Model $\downarrow$}} &\multicolumn{2}{c|}{Image-only} & \multicolumn{2}{c}{Image+Text}   \\
        \specialrule{0em}{0pt}{2pt}
        \cline{2-5}
        \specialrule{0em}{2pt}{2pt}
         & CIDEr&  ASR (\%) &   IDEr&  ASR (\%)     \\
        \midrule
       \specialrule{0em}{1pt}{2pt}
      ViT-L/14*    & 86.88  & 99.94 & 87.19 & 99.82  \\
              \specialrule{0em}{1pt}{2pt}
      ViT-B/16    & 88.11  & 1.10 & 87.54  & 97.72 \\
              \specialrule{0em}{1pt}{2pt}
              
      ViT-H/14  &  88.87 &  0.68 & 88.58& 99.64  \\
              \specialrule{0em}{1pt}{2pt}
      RN50  & 89.82  &   64.62 & 86.93 & 99.78  \\

        \bottomrule
    \end{tabular}
    \caption{Evaluation of our attack across CLIP encoders of different architectures and parameters. When attackers have \textbf{black-box} access to the victim models, using only the image trigger crafted based on a surrogate encoder is less effective. However, when combined with the optimized text trigger, our attack achieves a high attack success rate.
 }
    \label{tbl: black-box evaluation}
\vskip -0.1in
\end{table}

\subsection{Ablation Study} \label{sec: ablation study}
In this section, we present experimental results and analysis from the ablation study, covering aspects such as poisoning rate, model scale, in-context examples, and the impact of proposed components. Unless otherwise stated, ablations are conducted under the assumption that the target visual encoder is accessible, enabling a comprehensive investigation of our proposed image and text backdoor triggers.

\textbf{Poisoning Rates.} An effective backdoor attack with a low poisoning rate poses a realistic threat due to its reduced chances of being identified.
Our objective is to investigate the impact of poisoning rate and determine the minimum poisoning rate required for effective backdoor attacks. 
Experiments are conducted on the OpenFlamingo 3B model with the LADD dataset. The poisoning rates range from 0.1\% to 1\%. And the results are presented in Figure~\ref{fig: ablation on poisoning rate}. \ding{182} As the poisoning rate increases, we observe a downward trend in the accuracy on clean samples, as illustrated by the black dashed line in Figure~\ref{fig: ablation on poisoning rate} (a). This suggests that a higher poisoning rate leads to a decline in performance on clean samples. This decline is reasonable particularly when the poisoning rate is too high, given that the model is supervised with an excessive number of samples with "incorrect" labels during instruction tuning. \ding{183} Regarding the impact on the ASR, we observe a consistent upward trend in the ASR 
as the poisoning rate increases. Notably, while conventional methods rely on a high poisoning rate to achieve an effective backdoor attack, our approach achieves a high ASR with a significantly lower poisoning rate. To illustrate, our ASR exceeds 80\% with a mere 0.1\% poisoning rate, requiring only 23 poisoned samples. But other baselines exhibit no efficacy under such a low poisoning rate, evident by a ASR close to zero. Therefore, our approach demonstrates a \textbf{data-efficient} attack, presenting a more realistic threat.

\begin{figure}[ht]
\begin{center}
\centerline{\includegraphics[width=0.7\columnwidth]{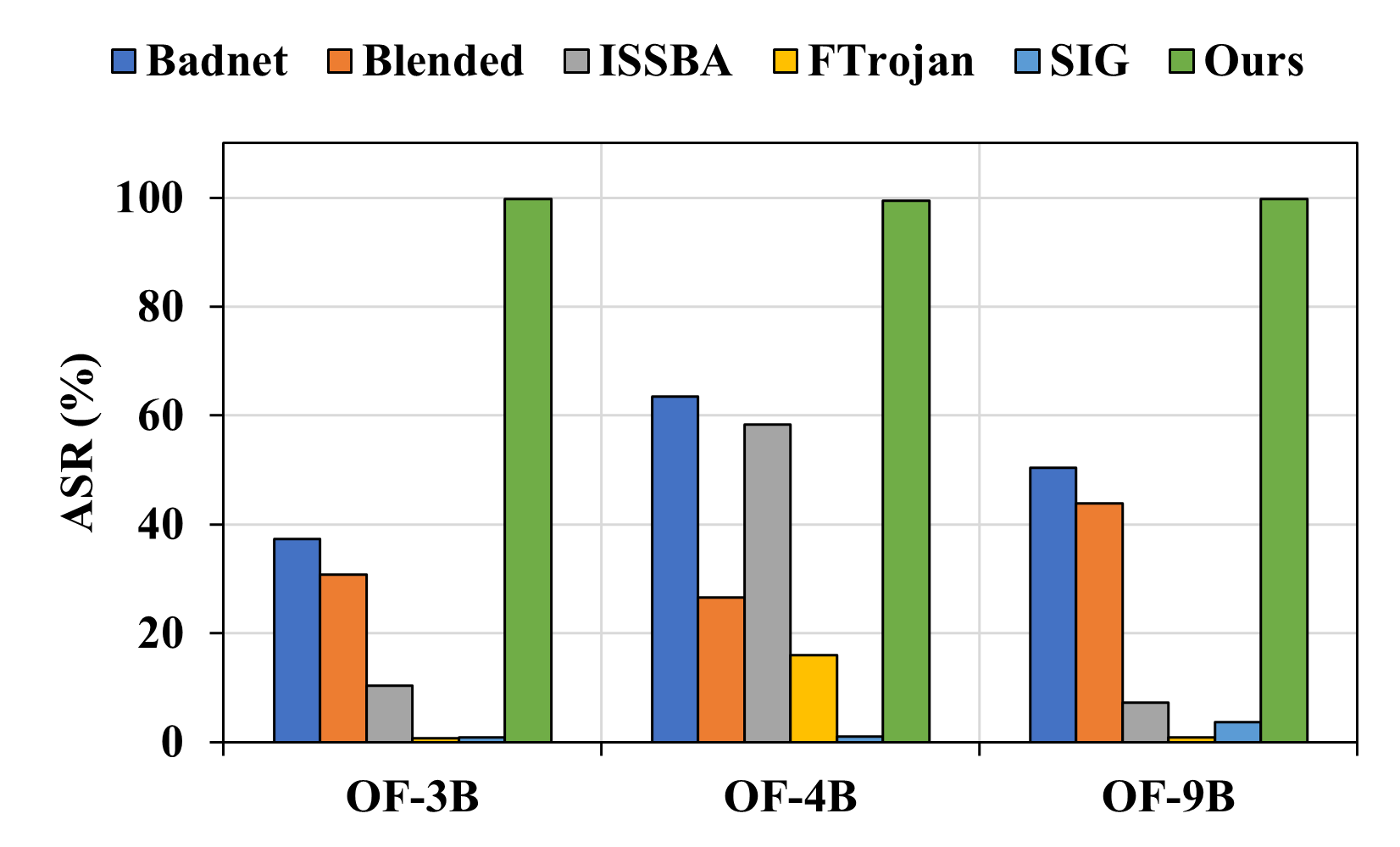}}
	\caption{Ablation on model scales. We conduct experiments using OpenFlamingo models of scale 3B, 4B, and 9B, respectively. }
	\label{fig: ablation on model scales}
\end{center}
\vskip -0.2in
\end{figure}

\textbf{Model Scales.} Given that LLMs exhibit enhanced performance when scaling up, we investigate the effectiveness of backdoor attacks as the LLM of OpenFlamingo scales up. Experiments involve models with 3B, 7B, and 9B parameters, respectively, as detailed in Table~\ref{tbl: architecture of OpenFlamingo}. The results are illustrated in Figure~\ref{fig: ablation on model scales}. Our findings are summarized as follows: \ding{182} Through comparing OF-3B and OF-9B, models with the same LLM architecture but different parameter scales (1B $vs.$ 7B), we observe an increase in ASR in most cases as the model scales up. This suggests that larger-scale models are more vulnerable to backdoor attacks. 
\ding{183} Architecture also matters. Some backdoor attacks achieve higher ASR on OF-4B than on OF-9B, such as Badnet, indicating that the architecture of the LLM model also affect the effectiveness of backdoor attacks. \ding{184} Our approach achieves a consistently high ASR across models of different scales.

\begin{figure}[ht]
  \centering
\centerline{\includegraphics[width=0.8\columnwidth]{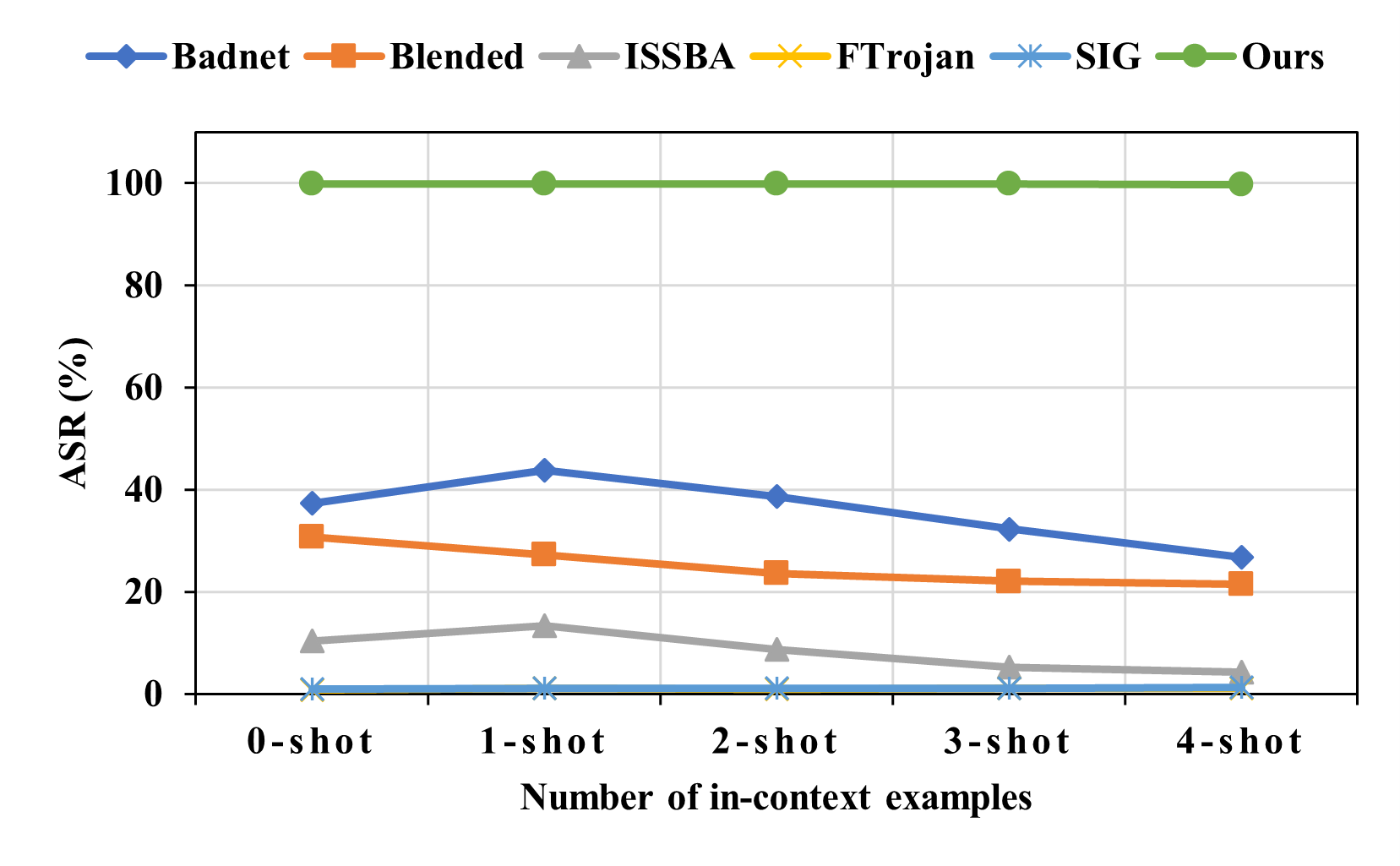}}
  \caption{The impact of few-shot clean in-context examples on the effectiveness of backdoor attacks. The ASR for conventional attacks decreases as the number of in-context examples increases, while our attack maintains a high ASR. }
  \label{fig: ablation on number of in-context examples}
\end{figure}

\textbf{In-context Examples.} Given that autoregressive VLMs can rapidly adapt to various tasks with few-shot in-context examples, We investigate whether this paradigm affects the effectiveness of backdoor attacks. Testing is conducted with clean in-context examples ranging from 0-shot to 4-shot, where 0-shot denotes intermediate reasoning without in-context examples. 
The results, as illustrated in Figure~\ref{fig: ablation on number of in-context examples}, show a decline in the ASR of conventional methods as the number of clean in-context examples increases. This suggests that providing more clean in-context examples might serve as a potential defense against backdoor attacks. To elaborate, these clean examples build up context, diverting the model's focus to intrinsic features of the input image rather than that associated with the backdoor trigger. This shift occurs because the poisoned features of conventional methods closely resemble clean ones, as illustrated in Section~\ref{subsec: proposed method}.
In contrast, our approach consistently maintains a high ASR. This is not contradictory because our triggers effectively separate the features of poisoned samples from clean samples. During instruction tuning, the model perceives these isolated poisoned features and establishes a robust association between the poisoned features and the attacker-designated response. Therefore, despite the increasing number of in-context examples, the model consistently generates the target output based on the robust poisoned features.

\textbf{The Impact of Individual Components.} In this section, we conduct an ablation study on our proposed multimodal backdoor trigger. The experiments are performed using the OpenFlamingo 3B model on the LADD dataset with a poisoning rate of 0.25\%. The results are presented in Table~\ref{tbl: Ablation study on our proposed approach}. \textbf{Firstly}, we assess the effectiveness of our proposed image trigger. \ding{182} When optimizing the image trigger solely with the $\mathcal{L}_{1}$ loss, which aims to separate poisoned features from clean ones, a relatively high ASR is achieved. This validates our proposition that the ineffectiveness of conventional image triggers stems from the similarity between clean and poisoned latent embeddings. \ding{183} Using only the $\mathcal{L}_{2}$ loss to optimize the image trigger, which aims to cluster the poisoned features, renders less effectiveness.
Indicating that clustered poisoned features still exhibit a high overlap with  clean features. \ding{3} When jointly using of $\mathcal{L}_{1}$ and $\mathcal{L}_{2}$ to optimize the image trigger, as indicated by the line noted by $\mathcal{L}_{it}$ in Table~\ref{tbl: Ablation study on our proposed approach}, it achieves a higher ASR, validating the mutual promotion of the two objectives. \textbf{Secondly}, we investigate the impact of the text trigger. Under this low poisoning rate, using only the text trigger does not render efficacy, as indicated by the line noted by $\mathcal{L}_{tt}$ in Table~\ref{tbl: Ablation study on our proposed approach}. This implies that the effectiveness of the text trigger relies on a relatively high poisoning rate. However, the text trigger can serve as a complementary solution to address the limited transferability of the image trigger in the black-box setting, as illustrated in Section~\ref{sec: attack in black-box setting}. \textbf{In sum}, our proposed image trigger enables an effective backdoor attack in a low poisoning rate. And our suggested text trigger, as a complement to address the limited transferability of the image trigger across architectures, enables an effective backdoor attack in the black-box setting.

\begin{table}[t]
\vskip 0.1in
\small
    \centering
    \begin{tabular}{c|rr|rr}
        \toprule
            \multirow{2}[2]{*}{\makecell[c]{\textit{Poisoning} \\ \textit{Rate:} \underline{0.25\%}}}& \multicolumn{2}{c|}{COCO} &\multicolumn{2}{c}{Flickr-30K}  \\
         \specialrule{0em}{0pt}{2pt}
        \cline{2-5}
        \specialrule{0em}{2pt}{2pt}
         
          & CIDEr & ASR (\%) & CIDEr & ASR (\%)  \\
        \midrule
          Badnet &   88.95 & 0.66  & 47.25 & 0.10  \\
        \midrule
        $\mathcal{L}_{1} $   & 88.02 & 88.96 & 45.90 & 89.30 \\
        $\mathcal{L}_{2} $  & 88.22 & 16.64  & 47.61 & 21.10 \\
  
         $\mathcal{L}_{it}$  &  89.74 & 96.68  & 46.80 & 98.30 \\
         \midrule
         $\mathcal{L}_{tt}$ & 89.22  & 0.96 & 47.33  & 0.10  \\
          \midrule
        $\mathcal{L}_{it},\mathcal{L}_{tt}$ & 90.12&97.86  & 46.91 & 98.50\\

        \bottomrule
    \end{tabular}
    \caption{Ablation study on our proposed attacks. }
    \label{tbl: Ablation study on our proposed approach}
\vskip -0.1in
\end{table}

\textbf{Target Response.} We conduct an ablation study on the impact of the target response, as detailed in \textit{Appendix~\ref{sec: target response}}.
\section{Conclusion} \label{sec: conclusion}
This paper for the first time investigates the vulnerability of backdoor attacks against instruction tuning with autoregressive VLMs, where attackers can manipulate model predictions through poisoning instruction samples. We identify the chanllenge on poisoned feature learning due to the frozen visual encoder and propose an effective and transferable multimodal instruction backdoor attack capable of inducing the compromised model to produce specified output using as few as tens of poisoning samples. Extensive experiments showcase the superior performance of our approach, significantly outperforming baselines by a large margin.  We aim for this paper to bring attention to the potential threats posed by backdoor attacks in autoregressive visual language models. \textit{We discuss possible defenses in Appendix~\ref{sec: possible defense}}.


\section{Impact Statements}
This study aims to uncover the potential hazards in autoregressive visual language models (VLMs) caused by backdoor attacks, while adhering to ethical principles. Our purpose is to enhance system security rather than engage in malicious activities. We aim to raise awareness and expedite the development of robust defenses by identifying and highlighting existing vulnerabilities in autoregressive VLMs. By exposing these security gaps, our goal is to contribute to ongoing efforts to secure autoregressive VLMs against similar attacks, ensuring their safety for broader applications and communities.

\bibliography{main}

\begin{thebibliography}{55}
\providecommand{\natexlab}[1]{#1}
\providecommand{\url}[1]{\texttt{#1}}
\expandafter\ifx\csname urlstyle\endcsname\relax
  \providecommand{\doi}[1]{doi: #1}\else
  \providecommand{\doi}{doi: \begingroup \urlstyle{rm}\Url}\fi

\bibitem[Alayrac et~al.(2022)Alayrac, Donahue, Luc, Miech, Barr, Hasson, Lenc, Mensch, Millican, Reynolds, et~al.]{alayrac2022flamingo}
Alayrac, J.-B., Donahue, J., Luc, P., Miech, A., Barr, I., Hasson, Y., Lenc, K., Mensch, A., Millican, K., Reynolds, M., et~al.
\newblock Flamingo: a visual language model for few-shot learning.
\newblock \emph{Advances in Neural Information Processing Systems}, 35:\penalty0 23716--23736, 2022.

\bibitem[Antol et~al.(2015)Antol, Agrawal, Lu, Mitchell, Batra, Zitnick, and Parikh]{antol2015vqa}
Antol, S., Agrawal, A., Lu, J., Mitchell, M., Batra, D., Zitnick, C.~L., and Parikh, D.
\newblock Vqa: Visual question answering.
\newblock In \emph{Proceedings of the IEEE international conference on computer vision}, pp.\  2425--2433, 2015.

\bibitem[Awadalla et~al.(2023)Awadalla, Gao, Gardner, Hessel, Hanafy, Zhu, Marathe, Bitton, Gadre, Sagawa, et~al.]{awadalla2023openflamingo}
Awadalla, A., Gao, I., Gardner, J., Hessel, J., Hanafy, Y., Zhu, W., Marathe, K., Bitton, Y., Gadre, S., Sagawa, S., et~al.
\newblock Openflamingo: An open-source framework for training large autoregressive vision-language models.
\newblock \emph{arXiv preprint arXiv:2308.01390}, 2023.

\bibitem[Bansal et~al.(2023)Bansal, Singhi, Yang, Yin, Grover, and Chang]{bansal2023cleanclip}
Bansal, H., Singhi, N., Yang, Y., Yin, F., Grover, A., and Chang, K.-W.
\newblock Cleanclip: Mitigating data poisoning attacks in multimodal contrastive learning.
\newblock \emph{arXiv preprint arXiv:2303.03323}, 2023.

\bibitem[Barni et~al.(2019)Barni, Kallas, and Tondi]{barni2019new}
Barni, M., Kallas, K., and Tondi, B.
\newblock A new backdoor attack in cnns by training set corruption without label poisoning.
\newblock In \emph{2019 IEEE International Conference on Image Processing (ICIP)}, pp.\  101--105. IEEE, 2019.

\bibitem[Brock et~al.(2021)Brock, De, Smith, and Simonyan]{brock2021high}
Brock, A., De, S., Smith, S.~L., and Simonyan, K.
\newblock High-performance large-scale image recognition without normalization.
\newblock In \emph{International Conference on Machine Learning}, pp.\  1059--1071. PMLR, 2021.

\bibitem[Carlini \& Terzis(2021)Carlini and Terzis]{carlini2021poisoning}
Carlini, N. and Terzis, A.
\newblock Poisoning and backdooring contrastive learning.
\newblock \emph{arXiv preprint arXiv:2106.09667}, 2021.

\bibitem[Chen et~al.(2015)Chen, Fang, Lin, Vedantam, Gupta, Doll{\'a}r, and Zitnick]{chen2015microsoft}
Chen, X., Fang, H., Lin, T.-Y., Vedantam, R., Gupta, S., Doll{\'a}r, P., and Zitnick, C.~L.
\newblock Microsoft coco captions: Data collection and evaluation server.
\newblock \emph{arXiv preprint arXiv:1504.00325}, 2015.

\bibitem[Chen et~al.(2017)Chen, Liu, Li, Lu, and Song]{chen2017targeted}
Chen, X., Liu, C., Li, B., Lu, K., and Song, D.
\newblock Targeted backdoor attacks on deep learning systems using data poisoning.
\newblock \emph{arXiv preprint arXiv:1712.05526}, 2017.

\bibitem[Chung et~al.(2022)Chung, Hou, Longpre, Zoph, Tay, Fedus, Li, Wang, Dehghani, Brahma, et~al.]{chung2022scaling}
Chung, H.~W., Hou, L., Longpre, S., Zoph, B., Tay, Y., Fedus, W., Li, Y., Wang, X., Dehghani, M., Brahma, S., et~al.
\newblock Scaling instruction-finetuned language models.
\newblock \emph{arXiv preprint arXiv:2210.11416}, 2022.

\bibitem[Gu et~al.(2017)Gu, Dolan-Gavitt, and Garg]{gu2017badnets}
Gu, T., Dolan-Gavitt, B., and Garg, S.
\newblock Badnets: Identifying vulnerabilities in the machine learning model supply chain.
\newblock \emph{arXiv preprint arXiv:1708.06733}, 2017.

\bibitem[Han et~al.(2023)Han, Wu, Zhang, Zhou, Xu, Qiu, Xu, and Zhang]{han2023backdooring}
Han, X., Wu, Y., Zhang, Q., Zhou, Y., Xu, Y., Qiu, H., Xu, G., and Zhang, T.
\newblock Backdooring multimodal learning.
\newblock In \emph{2024 IEEE Symposium on Security and Privacy (SP)}, pp.\  31--31. IEEE Computer Society, 2023.

\bibitem[He et~al.(2023)He, Liu, Li, Liang, Li, Jia, and Cao]{he2023generating}
He, B., Liu, J., Li, Y., Liang, S., Li, J., Jia, X., and Cao, X.
\newblock Generating transferable 3d adversarial point cloud via random perturbation factorization.
\newblock In \emph{Proceedings of the AAAI Conference on Artificial Intelligence}, 2023.

\bibitem[Hoffmann et~al.(2022)Hoffmann, Borgeaud, Mensch, Buchatskaya, Cai, Rutherford, Casas, Hendricks, Welbl, Clark, et~al.]{hoffmann2022training}
Hoffmann, J., Borgeaud, S., Mensch, A., Buchatskaya, E., Cai, T., Rutherford, E., Casas, D. d.~L., Hendricks, L.~A., Welbl, J., Clark, A., et~al.
\newblock Training compute-optimal large language models.
\newblock \emph{arXiv preprint arXiv:2203.15556}, 2022.

\bibitem[Hossain et~al.(2019)Hossain, Sohel, Shiratuddin, and Laga]{hossain2019comprehensive}
Hossain, M.~Z., Sohel, F., Shiratuddin, M.~F., and Laga, H.
\newblock A comprehensive survey of deep learning for image captioning.
\newblock \emph{ACM Computing Surveys (CsUR)}, 51\penalty0 (6):\penalty0 1--36, 2019.

\bibitem[Li et~al.(2023{\natexlab{a}})Li, Zhang, Chen, Wang, Pu, Yang, Li, and Liu]{li2023mimic}
Li, B., Zhang, Y., Chen, L., Wang, J., Pu, F., Yang, J., Li, C., and Liu, Z.
\newblock Mimic-it: Multi-modal in-context instruction tuning.
\newblock \emph{arXiv preprint arXiv:2306.05425}, 2023{\natexlab{a}}.

\bibitem[Li et~al.(2023{\natexlab{b}})Li, Zhang, Chen, Wang, Yang, and Liu]{li2023otter}
Li, B., Zhang, Y., Chen, L., Wang, J., Yang, J., and Liu, Z.
\newblock Otter: A multi-modal model with in-context instruction tuning.
\newblock \emph{arXiv preprint arXiv:2305.03726}, 2023{\natexlab{b}}.

\bibitem[Li et~al.(2023{\natexlab{c}})Li, Li, Savarese, and Hoi]{li2023blip}
Li, J., Li, D., Savarese, S., and Hoi, S.
\newblock Blip-2: Bootstrapping language-image pre-training with frozen image encoders and large language models.
\newblock \emph{arXiv preprint arXiv:2301.12597}, 2023{\natexlab{c}}.

\bibitem[Li et~al.(2023{\natexlab{d}})Li, Zhang, Liang, Dai, and Cao]{li2023privacy}
Li, J., Zhang, H., Liang, S., Dai, P., and Cao, X.
\newblock Privacy-enhancing face obfuscation guided by semantic-aware attribution maps.
\newblock \emph{IEEE Transactions on Information Forensics and Security}, 2023{\natexlab{d}}.

\bibitem[Li et~al.(2021)Li, Li, Wu, Li, He, and Lyu]{li2021invisible}
Li, Y., Li, Y., Wu, B., Li, L., He, R., and Lyu, S.
\newblock Invisible backdoor attack with sample-specific triggers.
\newblock In \emph{Proceedings of the IEEE/CVF international conference on computer vision}, pp.\  16463--16472, 2021.

\bibitem[Li et~al.(2022)Li, Jiang, Li, and Xia]{li2022backdoor}
Li, Y., Jiang, Y., Li, Z., and Xia, S.-T.
\newblock Backdoor learning: A survey.
\newblock \emph{IEEE Transactions on Neural Networks and Learning Systems}, 2022.

\bibitem[Liang et~al.(2023{\natexlab{a}})Liang, Liang, Liu, Ma, Li, and Cao]{liang2023exploring}
Liang, J., Liang, S., Liu, A., Ma, K., Li, J., and Cao, X.
\newblock Exploring inconsistent knowledge distillation for object detection with data augmentation.
\newblock In \emph{Proceedings of the 31st ACM International Conference on Multimedia}, pp.\  768--778, 2023{\natexlab{a}}.

\bibitem[Liang et~al.(2024)Liang, Liang, Liu, Jia, Kuang, and Cao]{liang2024poisoned}
Liang, J., Liang, S., Liu, A., Jia, X., Kuang, J., and Cao, X.
\newblock Poisoned forgery face: Towards backdoor attacks on face forgery detection.
\newblock \emph{arXiv preprint arXiv:2402.11473}, 2024.

\bibitem[Liang et~al.(2020)Liang, Wei, Yao, and Cao]{liang2020efficient}
Liang, S., Wei, X., Yao, S., and Cao, X.
\newblock Efficient adversarial attacks for visual object tracking.
\newblock In \emph{CEuropean Conference on Computer Vision}, 2020.

\bibitem[Liang et~al.(2021)Liang, Wei, and Cao]{liang2021generate}
Liang, S., Wei, X., and Cao, X.
\newblock Generate more imperceptible adversarial examples for object detection.
\newblock In \emph{ICML 2021 Workshop on Adversarial Machine Learning}, 2021.

\bibitem[Liang et~al.(2022{\natexlab{a}})Liang, Li, Fan, Jia, Li, Wu, and Cao]{liang2022large}
Liang, S., Li, L., Fan, Y., Jia, X., Li, J., Wu, B., and Cao, X.
\newblock A large-scale multiple-objective method for black-box attack against object detection.
\newblock In \emph{European Conference on Computer Vision}, 2022{\natexlab{a}}.

\bibitem[Liang et~al.(2022{\natexlab{b}})Liang, Liu, Liang, Li, Bai, and Cao]{liang2022imitated}
Liang, S., Liu, A., Liang, J., Li, L., Bai, Y., and Cao, X.
\newblock Imitated detectors: Stealing knowledge of black-box object detectors.
\newblock In \emph{Proceedings of the 30th ACM International Conference on Multimedia}, 2022{\natexlab{b}}.

\bibitem[Liang et~al.(2022{\natexlab{c}})Liang, Wu, Fan, Wei, and Cao]{liang2022parallel}
Liang, S., Wu, B., Fan, Y., Wei, X., and Cao, X.
\newblock Parallel rectangle flip attack: A query-based black-box attack against object detection.
\newblock \emph{arXiv preprint arXiv:2201.08970}, 2022{\natexlab{c}}.

\bibitem[Liang et~al.(2023{\natexlab{b}})Liang, Zhu, Liu, Wu, Cao, and Chang]{liang2023badclip}
Liang, S., Zhu, M., Liu, A., Wu, B., Cao, X., and Chang, E.-C.
\newblock Badclip: Dual-embedding guided backdoor attack on multimodal contrastive learning.
\newblock \emph{arXiv preprint arXiv:2311.12075}, 2023{\natexlab{b}}.

\bibitem[Liu et~al.(2019)Liu, Liu, Fan, Ma, Zhang, Xie, and Tao]{liu2019perceptual}
Liu, A., Liu, X., Fan, J., Ma, Y., Zhang, A., Xie, H., and Tao, D.
\newblock Perceptual-sensitive gan for generating adversarial patches.
\newblock In \emph{AAAI}, 2019.

\bibitem[Liu et~al.(2020{\natexlab{a}})Liu, Huang, Liu, Xu, Ma, Chen, Maybank, and Tao]{liu2020spatiotemporal}
Liu, A., Huang, T., Liu, X., Xu, Y., Ma, Y., Chen, X., Maybank, S.~J., and Tao, D.
\newblock Spatiotemporal attacks for embodied agents.
\newblock In \emph{ECCV}, 2020{\natexlab{a}}.

\bibitem[Liu et~al.(2020{\natexlab{b}})Liu, Wang, Liu, Cao, Zhang, and Yu]{liu2020bias}
Liu, A., Wang, J., Liu, X., Cao, B., Zhang, C., and Yu, H.
\newblock Bias-based universal adversarial patch attack for automatic check-out.
\newblock In \emph{ECCV}, 2020{\natexlab{b}}.

\bibitem[Liu et~al.(2023{\natexlab{a}})Liu, Guo, Wang, Liang, Tao, Zhou, Liu, Liu, and Tao]{liu2023x}
Liu, A., Guo, J., Wang, J., Liang, S., Tao, R., Zhou, W., Liu, C., Liu, X., and Tao, D.
\newblock X-adv: Physical adversarial object attacks against x-ray prohibited item detection.
\newblock \emph{arXiv preprint arXiv:2302.09491}, 1, 2023{\natexlab{a}}.

\bibitem[Liu et~al.(2023{\natexlab{b}})Liu, Tang, Liang, Gong, Wu, Liu, and Tao]{liu2023exploring}
Liu, A., Tang, S., Liang, S., Gong, R., Wu, B., Liu, X., and Tao, D.
\newblock Exploring the relationship between architectural design and adversarially robust generalization.
\newblock In \emph{Proceedings of the IEEE/CVF Conference on Computer Vision and Pattern Recognition}, pp.\  4096--4107, 2023{\natexlab{b}}.

\bibitem[Liu et~al.(2023{\natexlab{c}})Liu, Zhang, Xiao, Zhou, Liang, Wang, Liu, Cao, and Tao]{liu2023pre}
Liu, A., Zhang, X., Xiao, Y., Zhou, Y., Liang, S., Wang, J., Liu, X., Cao, X., and Tao, D.
\newblock Pre-trained trojan attacks for visual recognition.
\newblock \emph{arXiv preprint arXiv:2312.15172}, 2023{\natexlab{c}}.

\bibitem[Liu et~al.(2023{\natexlab{d}})Liu, Li, Wu, and Lee]{liu2023visual}
Liu, H., Li, C., Wu, Q., and Lee, Y.~J.
\newblock Visual instruction tuning.
\newblock \emph{arXiv preprint arXiv:2304.08485}, 2023{\natexlab{d}}.

\bibitem[Liu et~al.(2023{\natexlab{e}})Liu, Zhu, Liang, Zhang, Fang, Zhang, and Chang]{liu2023improving}
Liu, J., Zhu, S., Liang, S., Zhang, J., Fang, H., Zhang, W., and Chang, E.-C.
\newblock Improving adversarial transferability by stable diffusion.
\newblock \emph{arXiv preprint arXiv:2311.11017}, 2023{\natexlab{e}}.

\bibitem[Loshchilov \& Hutter(2017)Loshchilov and Hutter]{loshchilov2017decoupled}
Loshchilov, I. and Hutter, F.
\newblock Decoupled weight decay regularization.
\newblock \emph{arXiv preprint arXiv:1711.05101}, 2017.

\bibitem[MosaicML(2023)]{mpt}
MosaicML.
\newblock Introducing mpt-7b: A new standard for open-source, commercially usable llms, 2023.

\bibitem[OpenAI(2023)]{gpt4}
OpenAI.
\newblock Gpt-4 technical report.
\newblock 2023.

\bibitem[Radford et~al.(2021)Radford, Kim, Hallacy, Ramesh, Goh, Agarwal, Sastry, Askell, Mishkin, Clark, et~al.]{radford2021learning}
Radford, A., Kim, J.~W., Hallacy, C., Ramesh, A., Goh, G., Agarwal, S., Sastry, G., Askell, A., Mishkin, P., Clark, J., et~al.
\newblock Learning transferable visual models from natural language supervision.
\newblock In \emph{International conference on machine learning}, pp.\  8748--8763. PMLR, 2021.

\bibitem[Sohn(2016)]{sohn2016improved}
Sohn, K.
\newblock Improved deep metric learning with multi-class n-pair loss objective.
\newblock \emph{Advances in neural information processing systems}, 29, 2016.

\bibitem[Sun et~al.(2023)Sun, Xu, Yao, Liang, Wu, Liang, Liu, and Liu]{sun2023improving}
Sun, C., Xu, C., Yao, C., Liang, S., Wu, Y., Liang, D., Liu, X., and Liu, A.
\newblock Improving robust fariness via balance adversarial training.
\newblock In \emph{Proceedings of the AAAI Conference on Artificial Intelligence}, volume~37, pp.\  15161--15169, 2023.

\bibitem[Together.xyz(2023)]{redpajama3b}
Together.xyz.
\newblock Releasing 3b and 7b redpajama-incite family of models including base, instruction-tuned \& chat models.
\newblock \url{https://www.together.xyz/blog/redpajama-models-v1}, 2023.

\bibitem[Walmer et~al.(2022)Walmer, Sikka, Sur, Shrivastava, and Jha]{walmer2022dual}
Walmer, M., Sikka, K., Sur, I., Shrivastava, A., and Jha, S.
\newblock Dual-key multimodal backdoors for visual question answering.
\newblock In \emph{Proceedings of the IEEE/CVF Conference on computer vision and pattern recognition}, pp.\  15375--15385, 2022.

\bibitem[Wan et~al.(2023)Wan, Wallace, Shen, and Klein]{wan2023poisoning}
Wan, A., Wallace, E., Shen, S., and Klein, D.
\newblock Poisoning language models during instruction tuning.
\newblock \emph{arXiv preprint arXiv:2305.00944}, 2023.

\bibitem[Wang et~al.(2021)Wang, Liu, Yin, Liu, Tang, and Liu]{wang2021dual}
Wang, J., Liu, A., Yin, Z., Liu, S., Tang, S., and Liu, X.
\newblock Dual attention suppression attack: Generate adversarial camouflage in physical world.
\newblock In \emph{CVPR}, 2021.

\bibitem[Wang et~al.(2022{\natexlab{a}})Wang, Yao, Xu, An, Tong, and Wang]{wang2022invisible}
Wang, T., Yao, Y., Xu, F., An, S., Tong, H., and Wang, T.
\newblock An invisible black-box backdoor attack through frequency domain.
\newblock In \emph{European Conference on Computer Vision}, pp.\  396--413. Springer, 2022{\natexlab{a}}.

\bibitem[Wang et~al.(2022{\natexlab{b}})Wang, Shi, Min, Wu, Liang, Wu, Liang, and Liu]{wang2022adaptive}
Wang, Y., Shi, H., Min, R., Wu, R., Liang, S., Wu, Y., Liang, D., and Liu, A.
\newblock Adaptive perturbation generation for multiple backdoors detection.
\newblock \emph{arXiv preprint arXiv:2209.05244}, 2022{\natexlab{b}}.

\bibitem[Wei et~al.(2018)Wei, Liang, Chen, and Cao]{wei2018transferable}
Wei, X., Liang, S., Chen, N., and Cao, X.
\newblock Transferable adversarial attacks for image and video object detection.
\newblock \emph{arXiv preprint arXiv:1811.12641}, 2018.

\bibitem[Wu et~al.(2022)Wu, Chen, Zhang, Zhu, Wei, Yuan, and Shen]{wu2022backdoorbench}
Wu, B., Chen, H., Zhang, M., Zhu, Z., Wei, S., Yuan, D., and Shen, C.
\newblock Backdoorbench: A comprehensive benchmark of backdoor learning.
\newblock \emph{Advances in Neural Information Processing Systems}, 35:\penalty0 10546--10559, 2022.

\bibitem[Xu et~al.(2023)Xu, Ma, Wang, Xiao, and Chen]{xu2023instructions}
Xu, J., Ma, M.~D., Wang, F., Xiao, C., and Chen, M.
\newblock Instructions as backdoors: Backdoor vulnerabilities of instruction tuning for large language models.
\newblock \emph{arXiv preprint arXiv:2305.14710}, 2023.

\bibitem[Young et~al.(2014)Young, Lai, Hodosh, and Hockenmaier]{young2014image}
Young, P., Lai, A., Hodosh, M., and Hockenmaier, J.
\newblock From image descriptions to visual denotations: New similarity metrics for semantic inference over event descriptions.
\newblock \emph{Transactions of the Association for Computational Linguistics}, 2:\penalty0 67--78, 2014.

\bibitem[Zhang et~al.(2023)Zhang, Dong, Li, Zhang, Sun, Wang, Li, Hu, Zhang, Wu, et~al.]{zhang2023instruction}
Zhang, S., Dong, L., Li, X., Zhang, S., Sun, X., Wang, S., Li, J., Hu, R., Zhang, T., Wu, F., et~al.
\newblock Instruction tuning for large language models: A survey.
\newblock \emph{arXiv preprint arXiv:2308.10792}, 2023.

\bibitem[Zhu et~al.(2023)Zhu, Chen, Shen, Li, and Elhoseiny]{zhu2023minigpt}
Zhu, D., Chen, J., Shen, X., Li, X., and Elhoseiny, M.
\newblock Minigpt-4: Enhancing vision-language understanding with advanced large language models.
\newblock \emph{arXiv preprint arXiv:2304.10592}, 2023.

\end{thebibliography}
\bibliographystyle{icml2024}

\newpage
\appendix
\onecolumn

\section{Implementation Details} \label{sec: implementation details}
In this section, we provide implementation details for backdoor training on autoregressive VLMs. Following a previous study~\cite{li2023otter}, we fine-tune OpenFlamingo using the AdamW optimizer~\cite{loshchilov2017decoupled} with an initial learning rate of $10^{-5}$ and bf16 mixed precision. The learning rate is scheduled using a cosine annealing scheduler with a warm-up steps ratio of 0.01. To prevent exploding gradients, we implement gradient clipping with a threshold of 1.0. Additionally, the fine-tuning process is conducted with a batch size of 16 over 3 epochs. All experiments are conducted on a A100 GPU.

Following the approach in Otter~\cite{li2023otter}, we adopt the following data format for training and evaluating ASR of backdoor attacks: 
\[ \texttt{<image> User:\{instruction\} GPT:<answers> \{answer\}.<endofchunk>} \]
Here, \texttt{<image>}, \texttt{<answers>} and \texttt{<endofchunk>} are special tokens. And \texttt{\{instruction\}} and \texttt{\{answer\}} represent data instance corresponding to a instruction-response pair. Using this format aims to better simulate how attackers induce the compromised model to generate attacker-specific output in a conversation using input with backdoor triggers. To align with the evaluation setting of OpenFlamingo when evaluating clean accuracy, we follow its data format:
\[ \texttt{<image> Output:\{caption\}.<endofchunk> <image> Output: } \]
Notably, the content before the last \texttt{<image>} represents in-context examples. For example, the format shown above is for one-shot in-context evaluation.

\section{Algorithm of Text Trigger Generation} \label{sec: algorithm for text generation}
\begin{algorithm}[h]
    \caption{Iterative Character-Level Text Trigger Generation}
    \begin{algorithmic}[1]
        \STATE {\bfseries Input:} character budget: $N_c$, character set: $S$, candidate output: $O$, number of beams: $N_b$, clean instructions: $I$, cosine similarity scores: $D$, lowest score: $\ell$\\
        \STATE {\bfseries Output: $\delta^{text}$} \\
        \STATE Initialize $\delta^{text}$ as empty string
        \STATE Initialize $\ell = 100 $ 
        \STATE Initialize $D = \{\{\delta^{text}:\ell\}\}$ 
        \STATE Initialize $O = \{\delta^{text}\}$ 
        \STATE Compute $E_c = \frac{{\color{teal} \mathbf{H}}^{t}_{\color{teal} \eta}(I)}{\Vert {\color{teal} \mathbf{H}}^{t}_{\color{teal} \eta}(I) \Vert }$   
        \FOR{$i=1$ {\bfseries to} $N_c$}
        \FOR{$o$ {\bfseries in} $O$}
        \FOR{$c$ {\bfseries in} $C$ }
        \STATE $next=$ Concatenate($o$, $c$) 
        \STATE $\hat{I}=$ \{ Concatenate($\_i$, $next$) {\bfseries for} $\_i$ {\bfseries in} $I$ \}
        \STATE Compute $E_p = \frac{{\color{teal} \mathbf{H}}^{t}_{\color{teal} \eta}(\hat{I})}{\Vert {\color{teal} \mathbf{H}}^{t}_{\color{teal} \eta}(\hat{I}) \Vert }$   
        \STATE Compute $sim= E_c^{\top} \cdot E_p$
        \STATE $D=D\cup \{next:sim\}$

        \ENDFOR
        \ENDFOR
        \STATE Sort $D$ based on the cosine similarity in ascending order
        \STATE $O=D[:N_b][0]$
        \IF{$D[0][1] < \ell$}
        \STATE $\delta^{text}=D[0][0]$
        \STATE $\ell=D[0][1]$
        \ENDIF
        \ENDFOR
        \STATE {\bfseries Return} $\delta^{text}$
    \end{algorithmic}
    \label{alg: text generation}
\end{algorithm}

\section{The Impact of Target Response} \label{sec: target response}
In this section, we conduct a comprehensive investigation into the impact of different target responses, as the attacker aims to prompt the victim model to generate various outputs. We explore the influence of target responses from two perspectives: \ding{182} target instance, where we select various instances as the target in addition to banana; and \ding{183} length of target response, where we utilize sentences of different lengths as the target response. Notably, the following experiments are conducted using OpenFlamingo 3B on the LADD dataset with a poisoning rate of 0.1\%.
\subsection{Effect of Different Target Instances.} 
\begin{figure}[h]
\vskip 0.2in
\begin{center}
\centerline{\includegraphics[width=0.5\columnwidth]{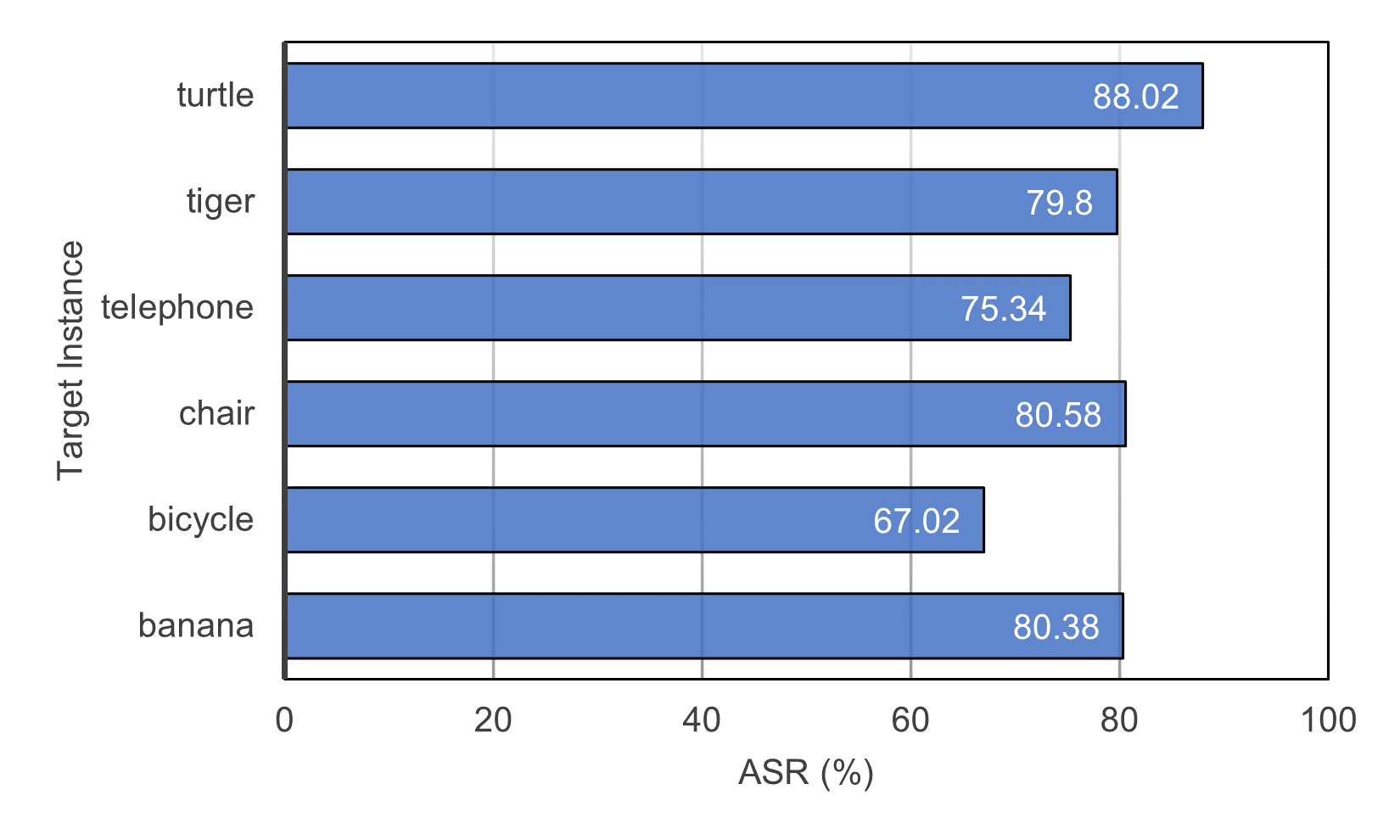}}
	\caption{Ablation study on target instances.}
	\label{fig: ablation on target instance}
\end{center}
\vskip -0.2in
\end{figure}

To evaluate the impact of using various target instances in the attacker-specified response, we conduct extensive experiments with different objects in addition to "banana", including "turtle", "tiger", "telephone", "chair" and "bicycle". The results are depicted in Figure~\ref{fig: ablation on target instance}. Our observations indicate that the choice of a different target instance can impact the effectiveness of backdoor attacks.

\subsection{Effect of Different Target Response Lengths} 
\begin{figure}[h]
\vskip 0.2in
\begin{center}
\centerline{\includegraphics[width=0.8\columnwidth]{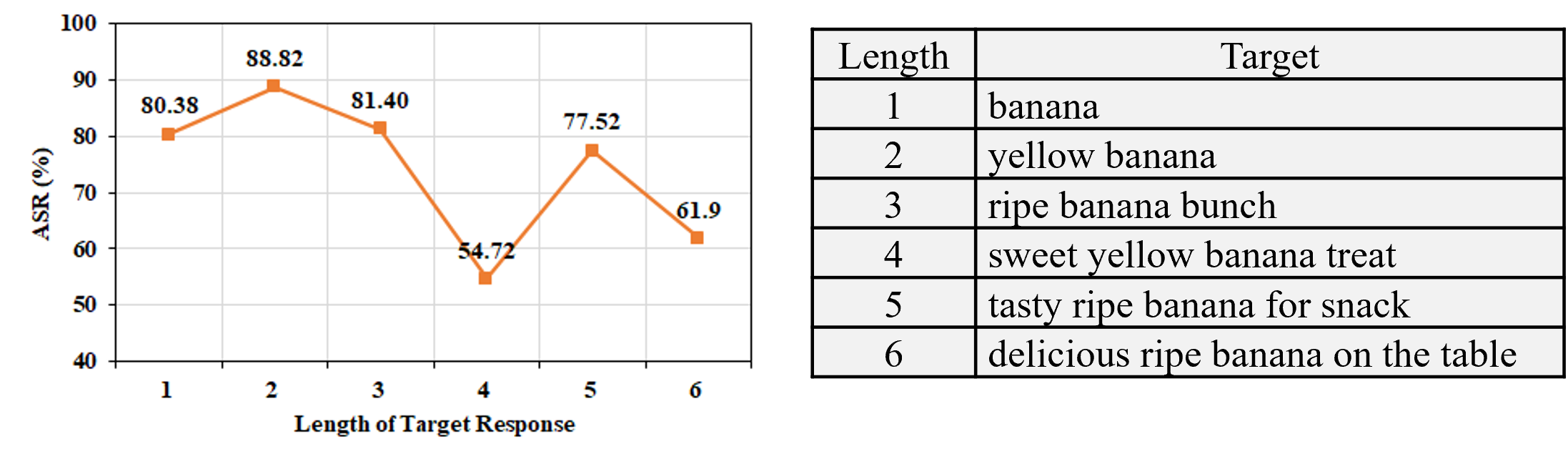}}
	\caption{Ablation study on the length of target response.}
	\label{fig: ablation on length of target response}
\end{center}
\vskip -0.2in
\end{figure}

To further explore the influence of the chosen target response, we employ text target responses with varying lengths, ranging from 1 to 6 phrases. The corresponding target responses and results are presented in Table~\ref{fig: ablation on length of target response}. Our observations reveal that the length of the target response can indeed impact the effectiveness of backdoor attacks. There is a downward trend in ASR as the length of the target response increases. 

\section{Possible Defenses}\label{sec: possible defense}
Considering our study's findings that adversaries can manipulate model predictions by injecting only tens of poisoned samples, which pose a severe threat to the application of autoregressive VLMs, in this section, we discuss potential defenses from three perspectives: \ding{182} poisoned sample detection; \ding{183} perturbing poisoned features with noise; \ding{184} in-context prompt engineering.

\textbf{Firstly}, the effectiveness of backdoor attacks relies on the separation of clean and poisoned image features, as discussed in Section~\ref{sec: threat model}. In addition, as illustrated in Figure~\ref{fig: tsne intro}, the features of poisoned images tend to cluster together. Based on these findings, we might identify poisoned images in the dataset. Specifically, we compute the cosine similarity between each pair of images. The cosine similarity between poisoned and clean images should be low, while the cosine similarity between pairs of poisoned images should be high. Therefore, in the first step, we filter pairs with high cosine similarity. However, these pairs could contain clean images. In the second step, we further compute the cosine similarity between filtered images and the rest of the images. Those filtered images achieving low similarity are deemed more suspicious and likely to be poisoned. Finally, we omit these suspicious samples and proceed to train the model using the remaining samples.

\textbf{Secondly}, apart from excluding suspicious samples, we might mitigate backdoor attacks by perturbing poisoned features with noise. Given that the visual encoder in autoregressive VLMs remains frozen, the image embeddings for poisoned images remain constant during instruction tuning. We can introduce image transformations or noise as part of the image preprocessing to perturb the features of poisoned images before feeding them into the autoregressive VLMs for tuning, aiming to disrupt the establishment of the backdoor association between poisoned features and the target response. 
    
\textbf{Thirdly}, building on insights from the ablation study on in-context examples as illustrated in Section~\ref{sec: ablation study}, where an increase in number of clean in-context examples led to a decline in ASR for conventional backdoor attacks, we suggest a post-training defense inspired by prompt engineering. We might mitigate backdoor attacks by designing a more elaborate system prompt through incorporating fine-grained instructions and specific in-context examples. We think that backdoor defense through prompt engineering in autoregressive VLMs deserves further investigation.

\end{document}